\documentclass[a4paper, 10pt, conference]{ieeeconf}      

\IEEEoverridecommandlockouts                              

\usepackage{graphicx} 
\usepackage{amsmath} 

\usepackage{amsthm}
\usepackage{listings}
\theoremstyle{definition}

\usepackage{amssymb}
\usepackage{pifont}
\usepackage{booktabs}
\usepackage{csvsimple}

\usepackage{caption}
\usepackage{subcaption}
\let\labelindent\relax
\usepackage{enumitem}
\usepackage{bm}
\usepackage{multirow}

\usepackage{algorithm}
\usepackage[noend]{algpseudocode}
\algnewcommand{\LineComment}[1]{\State \(\triangleright\) #1}
\usepackage[colorlinks=true,linkcolor=blue]{hyperref}
\usepackage[font=scriptsize,labelfont=bf]{caption}
\usepackage[dvipsnames]{xcolor}

\usepackage{caption}
\usepackage{subcaption}

\usepackage{cite}

\title{\LARGE \bf
Enabling Robots to Identify Missing Steps in Robot Tasks for Guided Learning from Demonstration*}

\author{Maximilian Diehl$^{1}$, Tathagata Chakraborti$^{2}$, and Karinne Ramirez-Amaro$^{1}$ 
\thanks{*This work is supported by Chalmers AI Research Centre (CHAIR).}
\thanks{$^{1}$Maximilian Diehl and Karinne Ramirez-Amaro. Faculty of Electrical Engineering, Chalmers University of Technology, SE-412 96 Gothenburg, Sweden.
        {\tt\small \{diehlm, karinne\}@chalmers.se}}%
\thanks{$^{2}$Tathagata Chakraborti is with IBM Research.}%
}

\theoremstyle{definition}
\newtheorem{defn}{Definition}

\begin{document}

\maketitle
\thispagestyle{empty}
\pagestyle{empty}

\begin{abstract}
Learning from Demonstration (LfD) systems are commonly used to teach robots new tasks by generating a set of skills from user-provided demonstrations. These skills can then be sequenced by planning algorithms to execute complex tasks. However, LfD systems typically require a full demonstration of the entire task, even when parts of it are already known to the robot. This limitation comes from the system's inability to recognize which sub-tasks are already familiar, leading to a repetitive and burdensome demonstration process for users.
In this paper, we introduce a new method for guided demonstrations that reduces this burden, by helping the robot to identify which parts of the task it already knows, considering the overall task goal and the robot's existing skills. In particular, through a combinatorial search, the method finds the smallest necessary change in the initial task conditions that allows the robot to solve the task with its current knowledge. This state is referred to as the ``excuse state.'' The human demonstrator is then only required to teach how to reach the excuse state (missing sub-task), rather than demonstrating the entire task. Empirical results and a pilot user study show that our method reduces demonstration time by 61\% and decreases the size of demonstrations by 72\%.
\end{abstract}

\begin{figure*}[!th]
\centering
\begin{subfigure}[b]{0.45\textwidth}
\centering
\includegraphics[width=\textwidth]{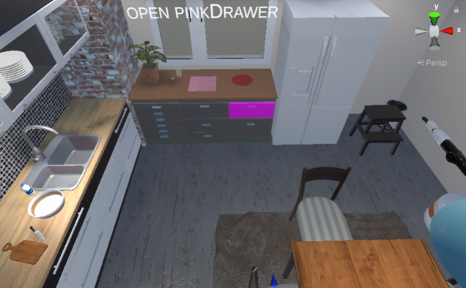}
\caption{Kitchen - I (showing excuse: \texttt{Open PinkDrawer})}
\label{fig:drawer_domain}
\includegraphics[width=\textwidth]{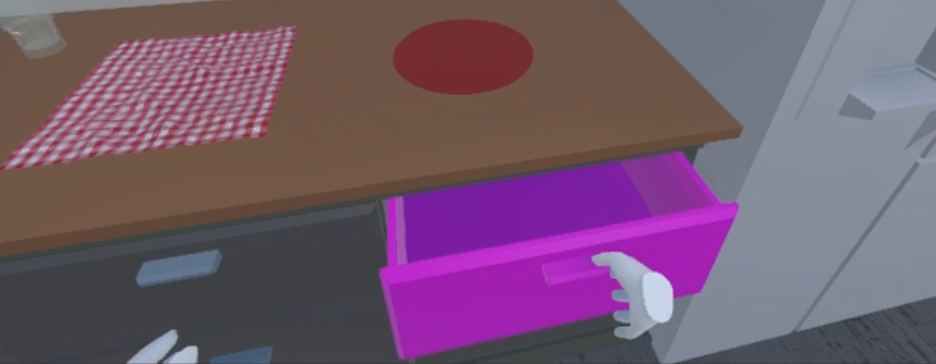}
\caption{Kitchen - I (ego perspective)}
\label{fig:drawer_domain_ego}
\end{subfigure}
\hfill
\begin{subfigure}[b]{0.45\textwidth}
\centering
\includegraphics[width=\textwidth]{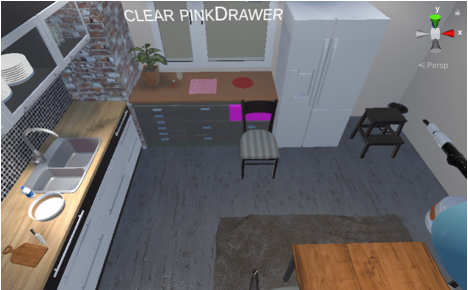}
\caption{Kitchen - II (showing excuse: \texttt{Clear PinkDrawer})}
\label{fig:drawer_chair_domain}
\includegraphics[width=\textwidth]{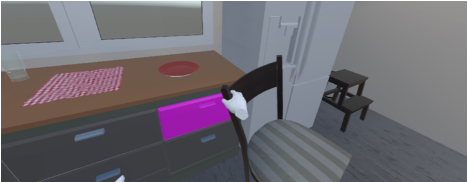}
\caption{Kitchen - II (ego perspective)}
\label{fig:drawer_chair_domain_ego}
\end{subfigure}
\hfill
\caption{
The Kitchen Domain (\ref{fig:drawer_domain}-\ref{fig:drawer_chair_domain_ego}) represents the class of HomeWorld domains~\cite{wisspeintner2009robocup} where the robot has to store a plate in a drawer, but initially does not know how to open a drawer (\ref{fig:drawer_domain}-\ref{fig:drawer_domain_ego}) and how to unblock the drawer from a chair (\ref{fig:drawer_chair_domain}-\ref{fig:drawer_chair_domain_ego}). Instead of having to demonstrate the full tasks, our proposed approach of guided demonstrations facilitates the teaching process, by instructing the human to only teach how to reach the automatically generated excuse states where the drawer is open (\ref{fig:drawer_domain}) and the drawer is clear (\ref{fig:drawer_chair_domain}). We directly communicate the obtained excuse states by displaying its symbolic state description in the VR environment used for the teaching process (\texttt{Open PinkDrawer} - Fig.~\ref{fig:drawer_domain} and \texttt{Clear PinkDrawer} - Fig.~\ref{fig:drawer_chair_domain}) as seen at the top of the respective images.}
\label{fig:domains}
\vspace{-5mm}
\end{figure*}

\section{Introduction}
Learning from demonstrations (LfD) is a widely used approach for teaching robots new tasks~\cite{argall2009survey}. It has been effectively applied to teach either individual skills in isolation or to create long-horizon task plans that involve a sequence of actions \cite{ekvall2008robot, konidaris2012robot}. Recent advances in LfD have focused on increasing flexibility and modularity by enriching the skills with a semantic skill description in the form of preconditions and effects \cite{steinmetz2019intuitive, diehl21, Zanchettin23}. 
This allows a symbolic planner to compute the required action sequence, leveraging both existing and newly learned skills.
A key advantage of this approach is that the robot can build a reusable skill library, which can be applied to future tasks.

While reusing skills across tasks is beneficial, the available skill set may sometimes be insufficient for completing a new task without additional demonstrations. For example, in Fig.~\ref{fig:drawer_domain}, the robot is tasked with cleaning up by placing plates into drawers. Previously, it learned to store a plate when the drawer was already open. However, in this task, the robot needs a new demonstration to learn how to open the drawer first. Without this missing sub-task, the robot cannot devise a successful plan to complete the goal. Unfortunately, robots typically lack the ability to reason about which sub-tasks are missing. As a result, a human must either demonstrate the entire task sequence—including parts already known—or manually identify and teach the missing skills. Re-teaching the complete task sequence can be tedious, repetitive, and ultimately reduce the user's willingness to continue teaching the robot. Requiring a deep understanding of the robot's skillset undermines the very purpose of learning from demonstrations, as such insight may not be readily available to the user.

To reduce the demonstration burden on the human instructor, we propose a novel method that identifies missing sub-tasks directly within the demonstration process, ensuring the user only needs to teach the missing components. We refer to this as a guided demonstration. Our method utilizes a combinatorial search, as described in~\cite{chakraborti-ijcai-2017-pl, caglar-aaai}, to find the nearest state, in the literature referred to as ``excuse state''~\cite{gobelbecker2010coming}, from which the robot's existing skills are sufficient to complete the task. By comparing the initial state with the excuse state, our approach identifies the smallest set of state changes needed to make an unsolvable task solvable.
Since our method relies on semantic state descriptions (e.g., \texttt{Inside Plate, Drawer} or \texttt{Clear Drawer}), these changes can be easily communicated to the user. For instance, as shown in Fig.~\ref{fig:domains}, we use a Virtual Reality (VR) LfD system where we display the required state changes (e.g., \texttt{Open PinkDrawer} for the cleaning task in Fig.~\ref{fig:drawer_domain}) directly in the environment. This reduces the demonstration burden from teaching the entire task of storing the plate to only teaching how to open the drawer~\footnote{Note that the used LfD system focuses on extracting high-level action descriptions from demonstrations rather than low-level execution data.}.

In this paper, we demonstrate how our method reduces the human effort required for teaching robots through demonstrations in task planning scenarios. Specifically, the contributions of this paper are: 1) the formalization of a new method for guided demonstrations that integrates the identification of missing sub-tasks into the teaching procedure using combinatorial search, and 2) a pilot user study that shows how guided demonstrations significantly reduce both the time and the number of demonstrations required, thereby improving the efficiency of LfD methods in expanding the robot’s ability to interact with its environment.

\section{Related Work}
Recent years have seen a large increase in learning from demonstration methods with contributions from different areas in robotics. 
The planning community has over three decades of work~\cite{callanan-icaps-2022-macq, ADL3}
centered around producing a symbolic domain model that satisfies a set of input
plans or ``traces''.
Approaches in this category cover a broad range of concerns such as partial 
observability~\cite{amir2008learning}, zero observability of states \cite{locm},
and noise~\cite{aman} -- but they all consider the observations
as a passive input with no considerations for a human demonstrator. 
Recent approaches use human demonstrations to obtain domain models.
In~\cite{ADL16, ADL12} kinesthetic teaching is used to generate semantic skill descriptions for reaching, grasping, pushing, and pulling.  In~\cite{Zanchettin23} industrial robotic tasks are taught. 
In~\cite{diehl21}, humans perform demonstrations in a Virtual Reality environment, a system recently extended to Multi-Agent scenarios~\cite{Diehl24}.
Even though these tools can be applied to teach robots new skills, they 
require complete task demonstrations or a user who decides what to teach.
These methods do not utilize prior knowledge to identify missing gaps in 
task plans. 

Some other works try to improve teaching efficiency by requiring the human only to teach the new task and utilize contextual knowledge in the form of ontologies~\cite{mitrevski21ontology} to connect the newly generated knowledge with the robot's prior task knowledge. However, similar to the works above, the robot itself is not capable of understanding what is missing. Another method~\cite{eibland-ras-23} stops the robot execution and asks for user demonstrations in case the robot reaches an unknown or abnormal state. However, in order to start the execution a robot plan had to be in place in the first place. We on the other hand tackle the problem when no task plan can be found in the first place. Other works try to reduce demonstration requirements by generalizing planning domains~\cite{tanneberg2023learning} but are still missing principled ways for the robot to detect missing gaps in unsolvable plans in the first place. Finally, some works such as~\cite{wang23} require feedback from a user when the robot tries to generalize an action from one object to another. This also reduces the demonstration effort, however, the robot is not able to detect and ask users if completely new actions are missing. 

Recently, large language models (LLMs) have been used to decompose and generate linguistic explanations of tasks~\cite{xie2023translating}. However, LLMs typically cannot produce low-level robot control strategies~\cite{dalal2024psl}, and learning policies for each sub-task through reinforcement learning can be expensive~\cite{dalal2024psl}. Consequently, these approaches, again, often rely on a predetermined set of robot actions from which plans are generated. This, however, poses a challenge when the existing set of actions is insufficient for the task at hand. This is the issue we address in our work. Recent research~\cite{caglar-aaai} has demonstrated the complementary strengths of LLMs versus combinatorial search for generating excuses for planning problems. In our work, while we use combinatorial search, our focus is not on the excuse generation process itself but on utilizing excuses to guide the user in identifying the missing steps that the robot needs to learn.

\section{The Guided Demonstration Task}
A Classical Planning Problem
\cite{geffner2013concise} is a tuple $\mathcal{M} = \langle D, I, G \rangle$ 
with domain $D = \langle F, A\rangle$ 
-- where $F$ is a set of fluents that define a state $s \subseteq F$, 
and $A$ is a set of actions -- and initial and goal states $I, G \subseteq F$. 
Action $a \in A$ is a tuple $\langle c_a, \textit{pre}(a), \textit{eff}^\pm(a)\rangle$ where $c_a$ is the cost, and $\textit{pre}(a), \textit{eff}^\pm(a) \subseteq F$ are the preconditions and add/delete effects, i.e. $\delta_{\mathcal{M}}(s, a) \models \bot \textit{ if } s\not\models \textit{pre}(a); \textit{ else } \delta_{\mathcal{M}}(s, a) \models s \cup \textit{eff}^+(a) \setminus \textit{eff}^-(a)$ where $\delta_{\mathcal{M}}(\cdot)$ is the transition function.

The cumulative transition function is a production over a sequence of 
actions, instead of a single action: $\delta_{\mathcal{M}}(s,\langle a_1, a_2, \ldots, a_n \rangle) = \delta_{\mathcal{M}}(\delta_{\mathcal{M}}(s, a_1),\langle a_2, \ldots, a_n \rangle)$.
The solution to $\mathcal{M}$ is a sequence of actions or a {\em plan} 
that transforms the initial state into a state that models the goal state:
$\pi = \langle a_1, a_2, \ldots, a_n \rangle$ such that $\delta_{\mathcal{M}}(I, \pi) \models G$. 
The cost of a plan $\pi$ is the sum of the costs of all the actions in 
it: $C(\pi) = \sum_{a\in\pi}c_a$ if $\delta_{\mathcal{M}}(I, \pi) \models G$; $\infty$ otherwise. 
Since we have two agents involved
-- the robot and the human demonstrator --
we will use subscripts $R$ and $H$, 
to refer to their respective domain artifacts
(e.g. $\delta_R$, $A_R$, and so on). 

An Unsolvable Planning Task
for the robot is one where there is no possible transition 
from the current state to the goal, i.e. 
$\not\exists \pi \textit{ such that } \delta_{R}(I_{R}, \pi) \models G_{R}$.

A human-in-the-loop can appear in several roles: 
as a collaborator, they can help the robot achieve its goal
as part of a joint effort.
They can also demonstrate how to achieve the goal and let the 
robot learn from observing them. In a demonstration scenario, the human does not have a goal but instead, demonstrates how to achieve the robot's goal with a plan using their own actions, and the robot
adopts one or more of the demonstrated actions for itself so it can achieve
the same goal:

\begin{defn}
\label{defn:demo}
\noindent{An (Unguided) Demonstration} for an unsolvable task, 
is a human plan $\pi_H$ that achieves the goal, wherein the robot can replicate parts 
of the demonstration to achieve the goal itself from the original initial state: 
$\delta_H(I_R, \pi) \models G_R$ such that $\delta_R(I_R, \mathbb{M}(\pi)) \models G_R$.
\end{defn}

To achieve this, a mapping function $\mathbb{M}$ translates the demonstrated human activity into robot-executable actions, known as an {\em embodiment mapping}~\cite{argall2009survey}. We assume this mapping is known and fixed, as our focus is on minimizing the demonstration effort. While demonstration learning is not purely confined to unsolvability (e.g. demonstrating a better way to achieve a goal), the problem of unsolvability and learning from demonstrations are intrinsically linked. Robots typically lack the ability to identify which sub-tasks are missing to complete a task. As a result, humans are often forced to either demonstrate the entire task, even when the robot already knows certain parts, or figure out which specific skills the robot is missing and teach those. This process of re-demonstrating the full task can be repetitive and frustrating, potentially discouraging users from continuing the teaching process. Furthermore, expecting users to have a detailed understanding of the robot's capabilities undermines the goal of learning from demonstrations, as this level of insight is often not readily available. To reduce the burden on the demonstrator, we therefore introduce the notion of excuses:

\begin{defn}
\label{defn:excuse}
\noindent{An Excuse} $\mathcal{E} = I \Delta I'$
is a change (symmetric difference) in the state of the world with a 
new state $I'$ that does not model the goal, such that
the unsolvable problem becomes solvable with the new state as the initial state:
$I_R \mapsto I'_R$, such that $I'_R \not\models G_R$ 
and $\exists \pi\ \delta_R(I'_R, \pi) \models G_R$.
\end{defn}

\begin{defn}
\label{defn:min-excuse}
\noindent{A minimal excuse} $\mathcal{E}_{min}$ is the smallest excuse 
that satisfies Definition \ref{defn:excuse}: 
$\mathcal{E}_{min}= \min_{I'}||I \Delta I'||$.
Unless otherwise mentioned, we will use an excuse to mean 
a minimal excuse.
\end{defn}
The excuse shown in Fig.~\ref{fig:drawer_domain} is a minimal excuse, with a size of 1. In contrast, a non-minimal excuse could involve additional conditions, such as having the plate inside the drawer and moving the chair away. Although this approach still leaves actions like closing the drawer necessary to fully achieve the goal and is thus a valid excuse, it is longer than the minimal excuse.
The concept of an excuse was originally defined in the seminal work by~\cite{gobelbecker2010coming} as a way to provide open-ended feedback for addressing issues in planning tasks. Since then, the notion has been expanded to include generic revisions of planning tasks~\cite{herzig2014revision} and has been used to debug planning models in various domain authoring tasks such as goal-oriented conversational agents, decision support systems, and intelligent tutoring systems~\cite{sreedharan-icaps-2020-d3w, grover-hci-2020-radar:-au, grover-icaps-2018-what-ca}.

In this paper, we propose a novel method for guided demonstrations that identifies missing sub-tasks by searching for a minimal excuse state. By focusing on demonstrating how to achieve this intermediate state rather than the entire original goal, our method significantly reduces the burden on the human demonstrator. We thus formalize our proposed guided demonstrations as follows: 
\vspace{5pt}
\begin{defn}
\label{defn:guided-demo}
\noindent{A Guided Demonstration} for an unsolvable task, 
is a human plan $\pi_H$ that demonstrates how to negate the excuse from
the current state (as opposed
to achieving the goal per Definition \ref{defn:demo}), 
wherein the robot can replicate parts 
of the demonstration to achieve the goal itself from the original state: 
$\delta_H(I_R, \pi_H) \models I_R + \mathcal{E}$ 
such that $\exists\pi\ \delta_R(I_R, \pi) \models G_R$ 
with $\exists a \in \pi$ and $a \in \mathbb{M}(\pi_H)$.
\end{defn}

\subsection{Measures for Guided Demonstration Effectiveness}
\label{sec:measures}
We compare guided and unguided demonstrations using the following objective metrics:
\subsubsection{Reduction in Demonstration Size}
\label{subsubsec:f1}
Our primary objective is to reduce the size of the demonstrations.
This can be measured by the relative sizes of guided versus unguided
demonstrations. Theoretically, the savings from a guided demonstration is the fractional length of the remaining human plan, from the excuse
state to the goal state, that they no longer had to demonstrate: 
$F_1 = |\pi'_H| \ / \ |\pi_H|$
where $\delta_H(I_R + \mathcal{E}, \pi'_H) \models G_R$ 
and $\delta_H(I_R, \pi_H) \models G_R$. Practically, we measure this by comparing the demonstration time between guided and unguided methods, and by comparing the planning action count produced by the devised LfD method. 

\subsubsection{Fraction of Useful Demonstrations}
\label{subsubsec:f2}
Even with guided demonstrations, there may be actions that the robot already knew about. Therefore, we can also measure the amount of new planning actions introduced by the guided demonstration, referred to as the \textit{fraction of useful demonstrations}, defined as: $||\{a\ |\ a \in \pi_H \textit{ such that } \mathbb{M}(a) \not\in A_R\}|| ~/~ |\pi_H|$.
For example, in Fig.~\ref{fig:drawer_chair_domain_ego}, all actions in the unguided demonstration after the drawer has been opened (such as placing the plate inside and closing it) will contribute to the savings in $F_1$. However, the actions leading up to opening the drawer (like picking up the plate and moving to the cabinet) are redundant, as the robot already knows how to perform them.

\subsubsection{Misdirected Guidance}
Depending on how the excuse is presented to the user, another failure mode might occur.
The excuse, as per its definition, ensures that the new state leads
to solvability. But this is not the only state where the excuse holds.
In other words, the demonstrator could solve the excuse in a way that leads to another unsolvable state by changing more facts than
what the excuse demanded: $\exists s, s \models I_R + \mathcal{E}$ 
but $\not\exists \pi, \delta_R(s, \pi) \models G_R$.

For example, in the Kitchen II domain (Fig.~\ref{fig:drawer_chair_domain}), the user might move the chair to a new position where the drawer is accessible, but now the chair obstructs the robot from reaching the plate. As a result, the robot may not find a successful plan for its task. Therefore, in addition to measuring the reduction in demonstration size and the reduction in redundant demonstrations, we can also evaluate the fraction of cases where guided demonstrations fail to enable the robot to find successful task plans due to misguided demonstrations. This serves as a third objective measure for the effectiveness of guided demonstrations. 

\subsubsection{Plan Solvability after Demonstration}
While misdirected guidance is one reason why a guided demonstration might fail to help the robot accomplish its task, other factors could be involved, such as the demonstrator misunderstanding the excuse and performing unhelpful actions. To generalize the measure of misdirected guidance, we assess whether the new domain, including the prior robot knowledge and the newly obtained domain (from either the excuse alone or the complete task demonstration), results in successful task plans:
$\exists\pi\ \delta_R(I_R, \pi) \models G_R$ 
with $\forall a \in \pi,\ a \in A_R \cup \mathbb{M}(\pi_H)$. 

\subsection{Caveats in Using Excuses as Guidance}
\label{subsec:caveats}
\subsubsection{Redundancy of the Excuse}
\label{subsubsec:redundancy}
An excuse can have some redundancy i.e. communicating a part of 
it might already achieve the desired effect: $\mathcal{E}' \subset \mathcal{E}_{min}$ 
such that Definition \ref{defn:guided-demo} still holds. This is not because 
$\mathcal{E}'$ is somehow a smaller 
excuse than $\mathcal{E}_{min}$ -- this is not possible and in fact, per
Definition \ref{defn:min-excuse}, $\mathcal{E}'$ is not even a valid excuse.
However, in the process of negating it, the demonstrator will achieve the 
same demonstration as they would have with a valid $\mathcal{E}$.

In the Kitchen II domain (Fig.~\ref{fig:drawer_chair_domain}), where the robot doesn't know how to open the drawer or remove the chair, a minimal excuse would involve clearing the chair and opening the drawer. However, simply asking for a demonstration on how to open the drawer would suffice, as the drawer can only be opened if it's clear. In this case, the human demonstrator would also need to move the chair, even though this action is not explicitly requested in the excuse (Fig.~\ref{fig:drawer_chair_domain_ego}). 

Unfortunately, predicting this situation before the demonstration is impossible because the robot does not know the human model, thus requiring the demonstration. Therefore, while this is an interesting caveat, we do not measure the redundancy of excuses in our guided demonstration setting.

\subsubsection{Preferred Excuse for Guidance}
\label{subsubsec:preferred-excuse}

Finally, there can be many different excuses for the same unsolvable task. 
An automated excuse generator that is optimizing for a minimality, 
will treat all excuses of the same size as equivalent.
This has been a stated issue \cite{zahedi-hri-2019-towards-u} in
existing model space search algorithms.
In the context of LfD, this has two implications: 
among logically equivalent excuses 1) some might
lead to smaller demonstrations -- this is akin to optimizing for the
usefulness measure in Section \ref{subsubsec:f2} instead of the size of an excuse; and
2) some might require an easier demonstration.
In Fig.~\ref{fig:drawer_domain}, where the goal is to store the plate in 
the drawer and close it, a request for either the plate to be in the drawer 
or just an open drawer are both minimal excuses but the latter will lead
to a shorter demonstration (Case 1 above).
Alternatively, imagine the path of your robot vacuum is blocked and you 
can move either a large sofa or a small chair to clear the path -- clearly, the latter
is easier to do and hence preferred (Case 2 above).

\subsection{Used Demonstration and Excuse Generation Methods}
VR has emerged as a powerful environment for demonstration learning due to
the relative ease in repeatability of experiments
and flexibility in constructing interesting environments to study \cite{walker-acm-2023-virtual-}. For learning planning domains from human demonstrations, we, therefore, use the Virtual Reality teaching system that was proposed in~\cite{diehl21}, with certain modifications, as described below.

\subsubsection{Automatic Domain Generation from Demonstrations}
\label{subsec:domainGen}

The first step of the system is activity recognition based on analyzing the hand movements of the demonstrators, as proposed in~\cite{BatesRIC17}. 
The demonstration is segmented and classified into high-level actions 
like \texttt{Reach, Take, Put, Stack or Idle}. 
The classification is based on a set of general rules in form of a C4.5 decision tree, which maps symbolic state variables like \texttt{inHand}, \texttt{actOn}, \texttt{handOpen}, \texttt{handMove} to the segmented activities (please refer to \cite{diehl21} for details on grounding of the used variables). 
Because of the generalizability of such semantic-based recognition methods, the same set of rules from~\cite{ramirez17AIJ} were re-used. The system utilizes the segmented actions to create new actions and their classification for a meaningful action name.
Additionally, the system also maps environment changes in the form of state variables such as \texttt{graspable}, \texttt{onTop}, \texttt{inside}, \texttt{clear} and \texttt{open} to the hand actions. 
Compared to~\cite{diehl21}, we updated how action-relevant fluents are selected. Previously, only fluents that changed during an action (e.g., \texttt{open} changing from False to True while opening a drawer) were included, ignoring static but critical fluents like \texttt{clear}. To address this, we now apply the \textit{nodes of interest} principle~\cite{Zanchettin23}, adding all fluents related to objects the user interacts with (e.g., the drawer), even if their values remain unchanged.

Note that we could have used different learning from demonstration systems such as kinesthetic LfD systems~\cite{eibland-ras-23} as long as their output has the form of a symbolic planning domain. However, considering that the explored environments are inspired by everyday human tasks such as opening kitchen drawers, performing the demonstration with their own hands (even if performed virtually) is more intuitive and doesn't require any knowledge or training about the robotic system.

\subsubsection{Automated Excuse Generation Algorithm}
\label{subsec:algo}

For excuse generation, we use the approach 
in \cite{chakraborti-ijcai-2017-pl} -- 
originally built for explanations in the form
of model reconciliation. It mimics the excuse generation 
algorithm in \cite{gobelbecker2010coming} but without a user
mental model. The approach performs combinatorial search
in the space of possible models $\mathcal{M}$ to find a model where the task is solvable. 
Starting from the initial state $I_R$ encoding the unsolvable task,
making one fluent change at a time to this task until a new state 
is reached where the task is solvable.
The excuse is the set of model edits (for this paper, confined only 
to the initial state of the planning problem being encoded) 
starting from the initial state of the search to the solvable state.
Since there are usually multiple alternative model edits that make the task solvable, we stop the search procedure when we find the first solution
(Fig.~\ref{fig:domains} shows some excuses communicated 
to the user). 

Our implementation is limited to PDDL~\cite{mcdermott1998pddl}, as that is the representation used in this setup. However, combinatorial search in model space applies broadly to various decision-making frameworks, including advanced representations, logic programs, and Markov Decision Processes~\cite{mirsky2019goal, kulkarni-iros-2020, son2021model}. Thus, excuse-guided demonstrations are, in principle, adaptable to these frameworks.

\section{User Study}
\label{subsec:userstudy}
 We hypothesize that guided demonstrations are more time efficient (H1) and 
reduce the demonstration size in the form of the number of demonstration steps (H2).

To evaluate these hypotheses, we conducted a pilot user study involving five participants. All participants were PhD students who had reported minimal experience with VR, ranging from 0 to 5 hours. Among them, three had experience with robots, one had basic knowledge, and one had no prior experience with robots.
Each participant was first introduced to the purpose of the study: helping robots to learn how to perform new tasks, by demonstrating them in our VR system. We also explained that they would only need to demonstrate how to reach the missing actions as specified by the excuse because the robot already has some prior knowledge. However, we did not further specify what this prior knowledge looks like to test the assumption that our method can work even if humans have no knowledge of the robot's current capabilities. We started the study with a short training phase, where the users were allowed to freely interact with blocks on a table so as to not induce any bias on the upcoming demonstrations. 
Each participant then encountered the two different scenarios in Fig.~\ref{fig:domains}. For each scenario, we asked for three demonstrations (excuse -- full task -- excuse). 
We asked them to demonstrate the excuse twice to investigate if knowing the full task goal has an impact on how the excuse is demonstrated. To present the participants with realistic excuses, 
we generated the robot's prior domain knowledge with the VR system 
based on demonstrations by one of the authors of this paper. The existing knowledge included actions enabling the robot to pick up a plate and place it in an open drawer. For the second excuse, actions for opening the drawer were added to the prior knowledge.
We chose the first minimal excuse generated
(\textit{open PinkDrawer} and \textit{clear PinkDrawer} 
for the two scenarios, respectively). 
The instructions for the complete task goal
\textit{RedPlate inside PinkDrawer} for both Kitchen domains. 
The excuses and task instructions were displayed visually inside the 
VR environment (Fig.~\ref{fig:domains}).

\section{Results}
\label{sec:results}
\subsection{Reduction in Time}

\begin{table}[]
\centering 
\small
 \begin{tabular}{ p{2cm} | p{0.6cm} | p{0.6cm}  | p{0.6cm}  | p{0.6cm}  | p{0.6cm}   }
 &  \textbf{P1} & \textbf{P2} & \textbf{P3} & \textbf{P4} & \textbf{P5}  \\ 
 \hline
\textbf{K I Excuse} & \textbf{3.71} & \textbf{8.77} & \textbf{4.38} & \textbf{8.19} & \textbf{4.4}  \\
 \hline
\textbf{K I Complete} & 9.64 & 30.65 & 10.87 & 15.57 & 13.94  \\
 \hline 
\textbf{K II Excuse} & \textbf{5.09} & \textbf{12.25} & \textbf{6.81} & \textbf{4.48} & \textbf{5.87}  \\
 \hline
\textbf{K II Complete} & 15 & 36.3 & 16.29 & 16.03 & 9.54  \\ 
 \hline 
\textbf{Ratio} & 0.36 & 0.314 & 0.412 & 0.4 & 0.44 \\
\end{tabular}
\caption{Demonstration time (in seconds) for every participant and scenario. On average the guided demonstration took 61\% less time than the complete demonstration. K = Kitchen Domains, Ratio = excuse-guided over complete demonstration time and bold numbers mark the smaller demonstration for each participant and task respectively comparing the excuse with the full demonstration.}
\label{tab:demonstrationTime}
\end{table}

Our evaluation of the guided demonstrations involves a comparison of the time taken by participants to complete the excuse and the full task demonstration
(Tab.~\ref{tab:demonstrationTime}). 
We observed that users required on average 61\% less time 
(mean of the time ratios reported in the bottom row)
to complete the guided demonstration compared to the full demonstration. This finding strongly supports our hypothesis H1 that guiding users through excuses reduces the time required for demonstrations.

\subsection{Theoretical Savings in Demonstration}
\begin{table}
\centering
\scriptsize
\begin{tabular}{@{}r|ccccc|ccccc@{}}
\textbf{} & \multicolumn{5}{c|}{\textbf{Kitchen I Scenario}} & \multicolumn{5}{c}{\textbf{Kitchen II Scenario}} \\ \midrule
\textbf{$\pi_H$} & \multicolumn{1}{c|}{\textbf{P}} & \multicolumn{1}{c|}{\textbf{O}} & \multicolumn{1}{c|}{\textbf{P}} & \multicolumn{1}{c|}{\textbf{P}} & \textbf{O} & \multicolumn{1}{c|}{\textbf{P}} & \multicolumn{1}{c|}{\textbf{O}} & \multicolumn{1}{c|}{\textbf{P}} & \multicolumn{1}{c|}{\textbf{P}} & \textbf{O} \\ \midrule
\textbf{} & \multicolumn{2}{c|}{\textbf{$\mathcal{E}_{min}$}} & \multicolumn{1}{c|}{\textbf{$\mathcal{E}$}} & \multicolumn{2}{c|}{\textbf{$G_R$}} & \multicolumn{2}{c|}{\textbf{$\mathcal{E}_{min}$}} & \multicolumn{1}{c|}{\textbf{$\mathcal{E}$}} & \multicolumn{2}{c}{\textbf{$G_R$}} \\ \midrule
\textbf{P1} & 2 & 3 & 7 & 9 & 9 & 1 & 2 & 9 & 11 & 9 \\ \midrule
\textbf{P2} & 3 & 7 & 8 & 10 & 25 & 4 & 4 & 10 & 12 & 19 \\ \midrule
\textbf{P3} & 2 & 4 & 7 & 9 & - & 1 & 2  & 9 & 11 & - \\ \midrule
\textbf{P4} & 2 & 3 & 7 & 9 & 12 & 1 & 2 & 9 & 11 & 10 \\ \midrule
\textbf{P5} & 2 & 3 & 7 & 9 & 7 & 1 & 2  & 9 & 11 & 7 
\end{tabular}
\caption{Predicted \textbf{P} (as obtained by an {\em automated planner}) versus observed \textbf{O} size of demonstrations $\pi_H$ for minimal $\mathcal{E}_{min}$, suboptimal excuse $\mathcal{E}$, and original task goal $G_R$. Demonstration size is measured in terms of number of planning actions, as obtained by the demonstration for \textbf{O} or the length of action plan as obtained by the planner based on the demonstrations for \textbf{P}. Note that we could not report values for the unguided demonstration from P3 (see columns \textbf{O} and $G_R$ for both kitchen scenarios) because P3 performed parallel hand activities (opening the drawer while picking up the plate), which were not supported by the version of the LfD tool used.}
\label{tab:theo}
\end{table}

While the reduction in demonstration time makes a compelling case for using excuses as guidance, we can also measure this reduction in terms of the demonstration size. We observed a reduction of 68\% and 77\% (average over the participant-specific ratios of $G_R$ over $\mathcal{E}_{min}$ for the observed (O) columns in Tab.~\ref{tab:theo}) for the two Kitchen domains, respectively. Note that we excluded P3 as the LfD system failed to generate a planning domain for the complete demonstration (column $G_R$ - O).

We also measure how closely the demonstrations resemble the theoretical prediction, to estimate how close to rational the demonstrators are over short demonstration periods.
In Tab.~\ref{tab:theo} we report the predicted plan length \textbf{P} of $\mathcal{E}_{min}$, obtained by an {\em automated planner}, with 
the plan length of a non-minimal excuse goal ($\mathcal{E}$) involving the plate being in
the drawer,
and the original goal ($G_R$). We can see that (aside from P2) the participants generally mimicked the theoretically expected demonstration length well and only demonstrated a few additional planning actions that were not included in the task plan. Additionally, we found that the minimal excuse $\mathcal{E}_{min}$ can lead to large savings in the demonstration length which underlines the importance of providing minimal excuses.

\subsection{Plan solvability after demonstrations}
We also evaluated if merging the prior robot domain with the planning actions that are obtained from the minimal excuse $\mathcal{E}$ demonstrations does in fact enable the robot to find a plan that satisfies the task goal $G_R$. 
We found that to be the case for all participants and both kitchen scenarios. 
This means that the excuse-guided demonstrations were sufficient 
for the robot to achieve its goal in all the scenarios (thereby confirming that the theory bears out in practice subjected to the users' 
interpretation of the guidance).

\subsection{Discussion}
\label{subsec:limits}
The user study showed our method's effectiveness, with a 61\% and 72\% reduction in time and demonstration size. Note that time savings may vary depending on the total plan length and the gap size. If the agent is missing a single action in a lengthy task (e.g., setting a table), we would expect even larger time savings. However, if the robot is missing multiple sub-tasks, the accomplished savings might be lower.

During the post-study interview, the participants expressed that the excuses were generally easy to comprehend. They rated their confidence in responding appropriately to the excuse with a demonstration as expected by the robot very positively, with a mean confidence score of 6.2 out of 7, where 7 signifies a very high level of confidence. Furthermore, even though we gave the users the chance to re-think how to address the excuse, after knowing the complete task goal, none of the users changed their excuse demonstration and all participants reported in the post-interview that they did not require knowledge of the overall task goal or prior robot knowledge, to understand how to perform a demonstration.

\section{Conclusion}
In this paper, we proposed a new method for guided demonstrations of robotic tasks that integrates the identification of missing sub-tasks into the teaching procedure using combinatorial search. Our user study showed that guided demonstration significantly reduces both the time and the number of demonstrations required, thereby improving the efficiency of LfD methods in expanding the robot’s ability to interact with its environment.
The current excuse-based guidance could face two issues: 1) misinterpretation, leading to incorrect or unhelpful demonstrations, and 2) lack of preference modeling, causing the robot to miss opportunities for more efficient or convenient demonstrations. Future work will explore ways to address these limitations, e.g., by applying LLMs to complement combinatorial search with latent world knowledge for generating more effective excuses~\cite{caglar-aaai}. 

\bibliography{root}

\begin{thebibliography}{10}
\providecommand{\url}[1]{#1}
\csname url@samestyle\endcsname
\providecommand{\newblock}{\relax}
\providecommand{\bibinfo}[2]{#2}
\providecommand{\BIBentrySTDinterwordspacing}{\spaceskip=0pt\relax}
\providecommand{\BIBentryALTinterwordstretchfactor}{4}
\providecommand{\BIBentryALTinterwordspacing}{\spaceskip=\fontdimen2\font plus
\BIBentryALTinterwordstretchfactor\fontdimen3\font minus \fontdimen4\font\relax}
\providecommand{\BIBforeignlanguage}[2]{{%
\expandafter\ifx\csname l@#1\endcsname\relax
\typeout{** WARNING: IEEEtran.bst: No hyphenation pattern has been}%
\typeout{** loaded for the language `#1'. Using the pattern for}%
\typeout{** the default language instead.}%
\else
\language=\csname l@#1\endcsname
\fi
#2}}
\providecommand{\BIBdecl}{\relax}
\BIBdecl

\bibitem{wisspeintner2009robocup}
T.~Wisspeintner, T.~Van Der~Zant, L.~Iocchi, and S.~Schiffer, ``{RoboCup@Home: Scientific Competition and Benchmarking for Domestic Service Robots},'' \emph{Interaction Studies}, 2009.

\bibitem{argall2009survey}
B.~D. Argall, S.~Chernova, M.~Veloso, and B.~Browning, ``{A Survey of Robot Learning from Demonstration},'' \emph{Robotics and Autonomous Systems}, 2009.

\bibitem{ekvall2008robot}
S.~Ekvall and D.~Kragic, ``{Robot Learning from Demonstration: A Task-level Planning Approach},'' \emph{International Journal of Advanced Robotic Systems}, 2008.

\bibitem{konidaris2012robot}
G.~Konidaris, S.~Kuindersma, R.~Grupen, and A.~Barto, ``{Robot Learning from Demonstration by Constructing Skill Trees},'' \emph{The International Journal of Robotics Research}, 2012.

\bibitem{steinmetz2019intuitive}
F.~Steinmetz, V.~Nitsch, and F.~Stulp, ``Intuitive task-level programming by demonstration through semantic skill recognition,'' \emph{IEEE Robotics and Automation Letters}, vol.~4, no.~4, 2019.

\bibitem{diehl21}
M.~Diehl, C.~Paxton, and K.~Ramirez-Amaro, ``{Automated Generation of Robotic Planning Domains from Observations},'' in \emph{IROS}, 2021.

\bibitem{Zanchettin23}
A.~M. Zanchettin, ``{Symbolic Representation of What Robots are Taught in One Demonstration},'' \emph{RAS}, 2023.

\bibitem{chakraborti-ijcai-2017-pl}
T.~Chakraborti, S.~Sreedharan, Y.~Zhang, and S.~Kambhampati, ``{Plan Explanations as Model Reconciliation: Moving Beyond Explanation as Soliloquy},'' \emph{Artificial Intelligence}, 2021.

\bibitem{caglar-aaai}
T.~Caglar, S.~Belhaj, T.~Chakraborti, M.~Katz, and S.~Sreedharan, ``{Can LLMs Fix Issues with Reasoning Models? Towards More Likely Models for AI Planning},'' in \emph{AAAI}, 2024.

\bibitem{gobelbecker2010coming}
M.~G{\"o}belbecker, T.~Keller, P.~Eyerich, M.~Brenner, and B.~Nebel, ``{Coming Up with Good Excuses: What To Do When No Plan Can be Found},'' in \emph{ICAPS}, 2010.

\bibitem{callanan-icaps-2022-macq}
E.~Callanan, R.~D. Venezia, V.~Armstrong, A.~Paredes, T.~Chakraborti, and C.~Muise, ``{MACQ: A Holistic View of Model Acquisition Techniques},'' in \emph{ICAPS Workshop on Knowledge Acquisition and Engineering (KEPS) and System Demonstration Track}, 2022.

\bibitem{ADL3}
A.~Arora, H.~Fiorino, D.~Pellier, M.~Métivier, and S.~Pesty, ``{A Review of Learning Planning Action Models},'' \emph{The Knowledge Engineering Review}, 2018.

\bibitem{amir2008learning}
E.~Amir and A.~Chang, ``{Learning Partially Observable Deterministic Action Models},'' \emph{Journal of Artificial Intelligence Research}, 2008.

\bibitem{locm}
S.~Cresswell and P.~Gregory, ``{Generalised Domain Model Acquisition from Action Traces},'' in \emph{ICAPS}, 2011.

\bibitem{aman}
H.~H. Zhuo and S.~Kambhampati, ``{Action-Model Acquisition from Noisy Plan Traces},'' in \emph{IJCAI}, 2013.

\bibitem{ADL16}
N.~Abdo, H.~Kretzschmar, L.~Spinello, and C.~Stachniss, ``{Learning Manipulation Actions from a Few Demonstrations},'' in \emph{ICRA}, 2013.

\bibitem{ADL12}
S.~R. {Ahmadzadeh}, A.~{Paikan}, F.~{Mastrogiovanni}, L.~{Natale}, P.~{Kormushev}, and D.~G. {Caldwell}, ``{Learning Symbolic Representations of Actions from Human Demonstrations},'' in \emph{ICRA}, 2015.

\bibitem{Diehl24}
M.~Diehl, I.~Zappa, A.~M. Zanchettin, and K.~Ramirez-Amaro, ``{Learning Robot Skills from Demonstration for Multi-Agent Planning},'' in \emph{CASE}, 2024.

\bibitem{mitrevski21ontology}
A.~Mitrevsk, P.~G. Plöger, and G.~Lakemeyer, ``Ontology-assisted generalisation of robot action execution knowledge,'' in \emph{IROS}, 2021.

\bibitem{eibland-ras-23}
T.~Eiband, J.~Liebl, C.~Willibald, and D.~Lee, ``{Online Task Segmentation by Merging Symbolic and Data-Driven Skill Recognition During Kinesthetic Teaching},'' \emph{Robotics and Autonomous Systems}, 2023.

\bibitem{tanneberg2023learning}
D.~Tanneberg and M.~Gienger, ``Learning type-generalized actions for symbolic planning,'' \emph{arXiv:2308.04867}, 2023.

\bibitem{wang23}
C.~Wang, A.~Belardinelli, S.~Hasler, T.~Stouraitis, D.~Tanneberg, and M.~Gienger, ``{Explainable Human-Robot Training and Cooperation with Augmented Reality},'' in \emph{CHI}, 2023, extended Abstract.

\bibitem{xie2023translating}
Y.~Xie, C.~Yu, T.~Zhu, J.~Bai, Z.~Gong, and H.~Soh, ``Translating natural language to planning goals with large-language models,'' 2023.

\bibitem{dalal2024psl}
M.~Dalal, T.~Chiruvolu, D.~Chaplot, and R.~Salakhutdinov, ``Plan-seq-learn: Language model guided rl for solving long horizon robotics tasks,'' in \emph{International Conference on Learning Representations}, 2024.

\bibitem{geffner2013concise}
H.~Geffner and B.~Bonet, ``A concise introduction to models and methods for automated planning,'' \emph{Synthesis Lectures on Artificial Intelligence and Machine Learning}, 2013.

\bibitem{herzig2014revision}
A.~Herzig, M.~V. de~Menezes, L.~N. De~Barros, and R.~Wassermann, ``{On the Revision of Planning Tasks},'' in \emph{ECAI}, 2014.

\bibitem{sreedharan-icaps-2020-d3w}
S.~Sreedharan, T.~Chakraborti, C.~Muise, Y.~Khazaeni, and S.~Kambhampati, ``{D3WA+ -- A Case Study of XAIP in a Model Acquisition Task for Dialogue Planning},'' in \emph{ICAPS}, 2020.

\bibitem{grover-hci-2020-radar:-au}
S.~Grover, S.~Sengupta, T.~Chakraborti, A.~P. Mishra, and S.~Kambhampati, ``{RADAR: Automated Task Planning for Proactive Decision Support},'' in \emph{HCI Journal}, 2020.

\bibitem{grover-icaps-2018-what-ca}
S.~Grover, T.~Chakraborti, and S.~Kambhampati, ``{What Can Automated Planning do for Intelligent Tutoring Systems?}'' in \emph{ICAPS Scheduling and Planning Applications Workshop}, 2018.

\bibitem{zahedi-hri-2019-towards-u}
Z.~Zahedi, A.~Olmo, T.~Chakraborti, S.~Sreedharan, and S.~Kambhampati, ``{Towards Understanding User Preferences for Explanation Types in Explanation as Model Reconciliation},'' in \emph{HRI Late Breaking Report}, 2019.

\bibitem{walker-acm-2023-virtual-}
M.~Walker, T.~Phung, T.~Chakraborti, T.~Williams, and D.~Szafir, ``{Virtual, Augmented, and Mixed Reality for Human-Robot Interaction: A Survey and Virtual Design Element Taxonomy},'' in \emph{ACM Transactions on Human Robot Interaction}, 2023.

\bibitem{BatesRIC17}
T.~Bates, K.~Ramirez-Amaro, T.~Inamura, and G.~Cheng, ``Online simultaneous learning and recognition of everyday activities from virtual reality performances,'' in \emph{IROS}, 2017.

\bibitem{ramirez17AIJ}
K.~Ramirez-Amaro, M.~Beetz, and G.~Cheng, ``{Transferring Skills to Humanoid Robots by Extracting Semantic Representations from Observations of Human Activities},'' \emph{Artificial Intelligence}, 2017.

\bibitem{mcdermott1998pddl}
D.~McDermott, M.~Ghallab, A.~Howe, C.~Knoblock, A.~Ram, M.~Veloso, D.~Weld, and D.~Wilkins, ``{PDDL -- The Planning Domain Definition Language},'' New Haven, CT: Yale Center for Computational Vision and Control, Tech. Rep. CVC TR98003/DCS TR1165, 1998.

\bibitem{mirsky2019goal}
R.~Mirsky, K.~Gal, R.~Stern, and M.~Kalech, ``{Goal and Plan Recognition Design for Plan Libraries},'' \emph{ACM Transactions on Intelligent Systems and Technology (TIST)}, 2019.

\bibitem{kulkarni-iros-2020}
A.~Kulkarni, S.~Sreedharan, S.~Keren, T.~Chakraborti, D.~E. Smith, and S.~Kambhampati, ``{Designing Environments Conducive to Interpretable Robot Behavior},'' in \emph{IROS}, 2020.

\bibitem{son2021model}
T.~C. Son, V.~Nguyen, S.~L. Vasileiou, and W.~Yeoh, ``{Model Reconciliation in Logic Programs},'' in \emph{Logics in Artificial Intelligence}, 2021.

\end{thebibliography}
\bibliographystyle{IEEEtran}

\end{document}